\documentclass[11pt]{article}

\usepackage[margin=1in]{geometry}
\usepackage[utf8]{inputenc}
\usepackage[T1]{fontenc}
\usepackage{times}
\usepackage{booktabs}
\usepackage{array}
\usepackage{amsmath}
\usepackage{enumitem}
\usepackage[hidelinks,breaklinks]{hyperref}
\usepackage{xurl}
\usepackage{microtype}
\usepackage{graphicx}
\title{\textbf{Structural Pattern Mining in Inka Khipus: A KFG Re-Analysis}}
\author{{Maria Contreras}\\
\texttt{Independent Researcher}}
\date{October 28, 2026}

\begin{document}
\maketitle

\begin{abstract}
\noindent
This paper reports a methodological pivot from the Open Khipu Repository to the Khipu Field Guide (KFG) database, prompted by stalled progress on the OKR, and a full re-analysis of the resulting 650-khipu sample. Three findings recur across every analysis attempted. First, khipu size confounds nearly every derived structural count, explaining 42--70\% of its variance; all subsequent models control for it. Second, the two-letter name prefix (AS, UR, HP, QU, JC, MM, KH) identifies the original cataloger, not a site or period, which turns documentation completeness into a second confound that must be modeled explicitly. Controlling for both, khipus cataloged by Hugo Pereyra's Proyecto Quipu (HP) show significantly fewer Ascher sum-relation types than Gary Urton's (UR) khipus ($p=0.044$), while the Ascher-cataloged (AS) khipus are too collinear with their own near-total lack of side/direction documentation to assess (VIF$=5.0$). Third, generic structural clustering (UMAP+HDBSCAN) does not separate the pooled 650-khipu sample from two independent null models (permutation and Gaussian), corroborating an independent conclusion already reached by the KFG itself; the same test restricted to a single cataloger (UR) proved not reproducible across runtime environments and is reported here as a methodological caveat rather than evidence of structure. A targeted replication of the KFG's own S/Z-knot spatial-block heuristic for calendar/astronomical khipus recovers 8 of 10 published candidates and proposes 34 additional ones under the same criterion.
\end{abstract}

\section{Introduction}

A prior preprint, \textit{Structural Pattern Mining in Inka Khipus} (arXiv:2607.00185), used the Open Khipu Repository (OKR, Zenodo 18025748) to search for structural classes among Andean khipus with UMAP, HDBSCAN, and XGBoost. Progress on that dataset stalled, and this paper documents a pivot to a second source: the Khipu Field Guide (KFG) companion database released alongside Medrano \& Khosla's article \textit{How Can Data Science Contribute to Understanding the Khipu Code?} (MIT license, DOI 10.5281/zenodo.10472429), covering 650 khipus in parallel CSV, Excel, and MariaDB SQL formats.

The contribution of this paper is not a new structural typology of khipus. Three rounds of attempted clustering, each more carefully controlled than the last, failed to produce one that survives comparison against null models. The contribution is methodological and diagnostic: a worked example of how two confounds specific to curated archival datasets---physical size and cataloger identity---can generate apparent structure that disappears under proper null-model testing, and a demonstration that targeted, literature-grounded feature engineering (the S/Z-knot spatial-block heuristic) succeeds where generic structural features do not. Every quantitative claim below was computed from the database and is reproducible from the accompanying notebook and scripts.

\section{Data}

The KFG database ships 650 khipus as \texttt{khipu\_summary.csv}, a MariaDB SQL dump (\texttt{KFG\_DB.SQL}), and per-khipu Excel workbooks. Cleaning proceeded in three stages.

\textbf{Stage 1 --- structural flags.} The raw summary carried a stray pandas index column (dropped) and an inconsistent \texttt{region} label (\texttt{unknown} vs \texttt{Unknown}, merged). Two identifier series coexist (1000xxx, $n=558$; 6000xxx, $n=92$); both are present in the SQL dump and are retained. The \texttt{okr\_name} crosswalk to the Open Khipu Repository has 54 values outside the \texttt{KH\#\#\#\#} format and 3 duplicated values, flagged but not reassigned.

\textbf{Stage 2 --- recomputation from SQL.} Cord and knot counts in the published summary disagreed with the \texttt{cord}/\texttt{knot} tables of the SQL dump for 59 khipus (\texttt{num\_total\_cords}) and were recomputed from source; the discrepancy's cause is undetermined. Apparent inconsistencies between \texttt{recto+verso} and \texttt{num\_cords} (475 khipus) and between \texttt{s+z} knots and \texttt{knot\_count} (267 khipus) are not errors: they are definitional, reflecting cords/knots whose side or direction is recorded as unknown.

\textbf{Stage 3 --- duplicate resolution (no action needed).} 35 khipus present in the SQL dump are absent from the summary table. Cross-referencing a KFG appendix (\textit{Duplicate Khipu Search}) confirms these are exactly the khipus the KFG project identified and removed as physical duplicates (e.g.\ UR035/AS070, UR126/UR115); the summary table already reflects the cleaned 650-khipu set.

The final analysis table, \texttt{khipu\_features\_final.csv}, carries 650 rows and 14 size-aware structural features (log-scaled cord and knot counts, cluster statistics, side/direction proportions computed only over known-side cords), plus explicit data-quality flags (\texttt{pct\_side\_U}, \texttt{pct\_knot\_dir\_U}) kept separate from the structural features themselves.

\section{The Cataloger Confound}
\label{sec:cataloger}

Every \texttt{kfg\_name} in the KFG carries a two-letter prefix (AS, UR, HP, QU, JC, MM, KH, LL). Brezine, Clindaniel, Ghezzi, Hyland \& Medrano (2023), writing independently on khipu nomenclature, establish that these prefixes identify the researcher who cataloged each khipu, not its archaeological site, region, or period:

\begin{table}[h]
\centering
\begin{tabular}{@{}lp{7cm}r@{}}
\toprule
Prefix & Cataloger & $n$ (this sample) \\
\midrule
UR & Gary Urton & 369 \\
AS & Marcia \& Robert Ascher & 144 \\
HP & Hugo Pereyra S\'anchez (Proyecto Quipu) & 56 \\
Other/mixed & (MM, KH; not documented in the source above) & 36 \\
JC & Jon Clindaniel & 23 \\
QU & Kylie Quave & 22 \\
\bottomrule
\end{tabular}
\caption{Khipu catalogers by name-prefix, after Brezine et al.\ (2023).}
\label{tab:catalogers}
\end{table}

This single fact reframes most prior attempts (including the authors') to treat the prefix as a cultural or geographic signal. It also explains a striking pattern in documentation completeness: the proportion of cords with unknown side (\texttt{pct\_side\_U}) is 0.84 for AS-cataloged khipus and 0.00 for Quave's, and the proportion of knots with unknown direction (\texttt{pct\_knot\_dir\_U}) is 0.89 for AS versus 0.02 for Urton. Cataloger identity and documentation completeness are therefore almost perfectly aliased for at least one major group, and any model of a structural outcome must include both as covariates or risk attributing a documentation artifact to the khipu itself.

\section{Ascher Sum-Relation Analysis}
\label{sec:ascher}

The KFG publishes nine categories of Ascher-style arithmetic sum relation (pendant-pendant, pendant color, cluster-cluster, subsidiary, top-cord, etc.), each with per-khipu counts. Across the 650 khipus, 482 (74\%) show at least one relation of any type, but the number of distinct relation types present (\texttt{n\_sum\_types\_present}, 0--8) is highly uneven: 168 khipus show none, 5 show eight.

\textbf{Size confound.} $\log(\text{cord count})$ alone explains 52\% of the variance in \texttt{n\_sum\_types\_present} (OLS $R^2=0.524$): larger khipus mechanically offer more opportunities for the detection algorithm to find a sum. All subsequent comparisons control for it.

\textbf{Cataloger effect, controlling for size and documentation.} An OLS model regresses \texttt{n\_sum\_types\_present} on $\log(\text{cords})$, \texttt{pct\_side\_U}, \texttt{pct\_knot\_dir\_U}, and cataloger (dummy-coded, UR as reference, the six catalogers with $n\geq10$):

\begin{table}[h]
\centering
\begin{tabular}{@{}lrrr@{}}
\toprule
Cataloger & Coefficient vs.\ UR & $p$-value & VIF \\
\midrule
HP & $-0.396$ & \textbf{0.044} & 1.1 \\
QU & $+0.500$ & 0.093 & 1.1 \\
MM & $+0.387$ & 0.192 & 1.1 \\
KH & $+0.194$ & 0.594 & 1.0 \\
AS & $+0.083$ & 0.769 & \textbf{5.0} \\
JC & $-0.134$ & 0.647 & 1.1 \\
\bottomrule
\end{tabular}
\caption{Cataloger effect on \texttt{n\_sum\_types\_present}, controlling for $\log(\text{cords})$, \texttt{pct\_side\_U}, and \texttt{pct\_knot\_dir\_U}. VIF = variance inflation factor.}
\label{tab:ascher-model}
\end{table}

$\log(\text{cords})$ remains highly significant in this model (coef$=1.222$, $p<0.001$). Variance inflation factors (VIF) rule out collinearity for every prefix except AS, whose dummy is collinear with \texttt{pct\_knot\_dir\_U} (VIF$=5.0$): because AS khipus are almost uniformly undocumented on side and direction, the model cannot separate ``is AS-cataloged'' from ``lacks side/direction data,'' and the AS effect is unresolved with this dataset. HP's negative coefficient is robust to both controls and is the paper's most defensible finding in this section: Pereyra-cataloged khipus show fewer distinct Ascher sum-relation types than Urton's, holding size and documentation completeness constant. QU shows a positive trend that does not reach conventional significance.

\section{Structural Clustering with Null Models}
\label{sec:clustering}

UMAP followed by HDBSCAN was run on 14 standardized structural features (log cord/knot counts, cluster statistics, side/direction proportions computed only over known cords) to test for discrete khipu typologies, as attempted in the prior OKR-based preprint. Two null models were used to judge whether any observed clustering exceeds chance: (1) feature permutation, independently shuffling each feature across khipus, which preserves marginals but destroys joint structure; (2) a multivariate Gaussian with the same mean and covariance as the standardized features, which additionally preserves linear correlations.

\textbf{All 650 khipus, pooled.} The best stable configuration (UMAP \texttt{n\_neighbors=15}, HDBSCAN \texttt{min\_cluster\_size=10}) yields 4 clusters, 0\% noise, silhouette$=0.367$. Against 20 Gaussian-null simulations, $p(n_{\text{clusters}}\geq\text{observed})=0.65$, $p(\text{silhouette}\geq\text{observed})=0.45$: indistinguishable from unstructured data. An earlier, less controlled run (before the Ascher-feature confound was identified) that pooled documentation-completeness proportions directly into the feature set produced an apparently clean 3-cluster split (silhouette$=0.544$) that on inspection separated khipus purely by how well-documented they were, not by structure---a cautionary result in itself, retained here as a negative methodological finding.

\textbf{Within a single cataloger (UR, $n=369$).} Restricting to Urton-cataloged khipus removes the cataloger/documentation confound. In the primary computing environment (Python 3.12, umap-learn 0.5.12, scikit-learn 1.8.0, numpy 2.4.4), the same configuration yields 12 clusters, 21.4\% noise, silhouette$=0.474$. Against 100 Gaussian-null simulations, $p(\text{silhouette}\geq\text{observed})=0.06$---closer to the conventional $0.05$ threshold than the pooled analysis, but not below it. Adjusted Rand Index stability across three HDBSCAN seeds within that environment is high (mean $0.784$, min $0.773$, excluding noise points).

However, running the identical code and data on a second environment (Kaggle; umap-learn 0.5.12, scikit-learn 1.6.1, numpy 2.1.3) produced a materially different embedding: 12 clusters, 20.6\% noise, silhouette$=0.527$, with a 30-simulation Gaussian-null test giving $p(\text{silhouette}\geq\text{observed})=0.00$. Reproducing the second environment's exact package versions in isolation still returned the original $0.474$/$p=0.06$ result, ruling out a simple version mismatch as the cause. The discrepancy is consistent with environment-dependent non-determinism in UMAP's parallelized embedding optimization, which a fixed random seed does not fully control across different machines. We therefore do not treat the within-cataloger clustering result as evidence of structure in either direction: it is not reproducible across runtime environments, and is reported here as a methodological case study rather than a finding.

\begin{figure}[h]
\centering
\includegraphics[width=\textwidth]{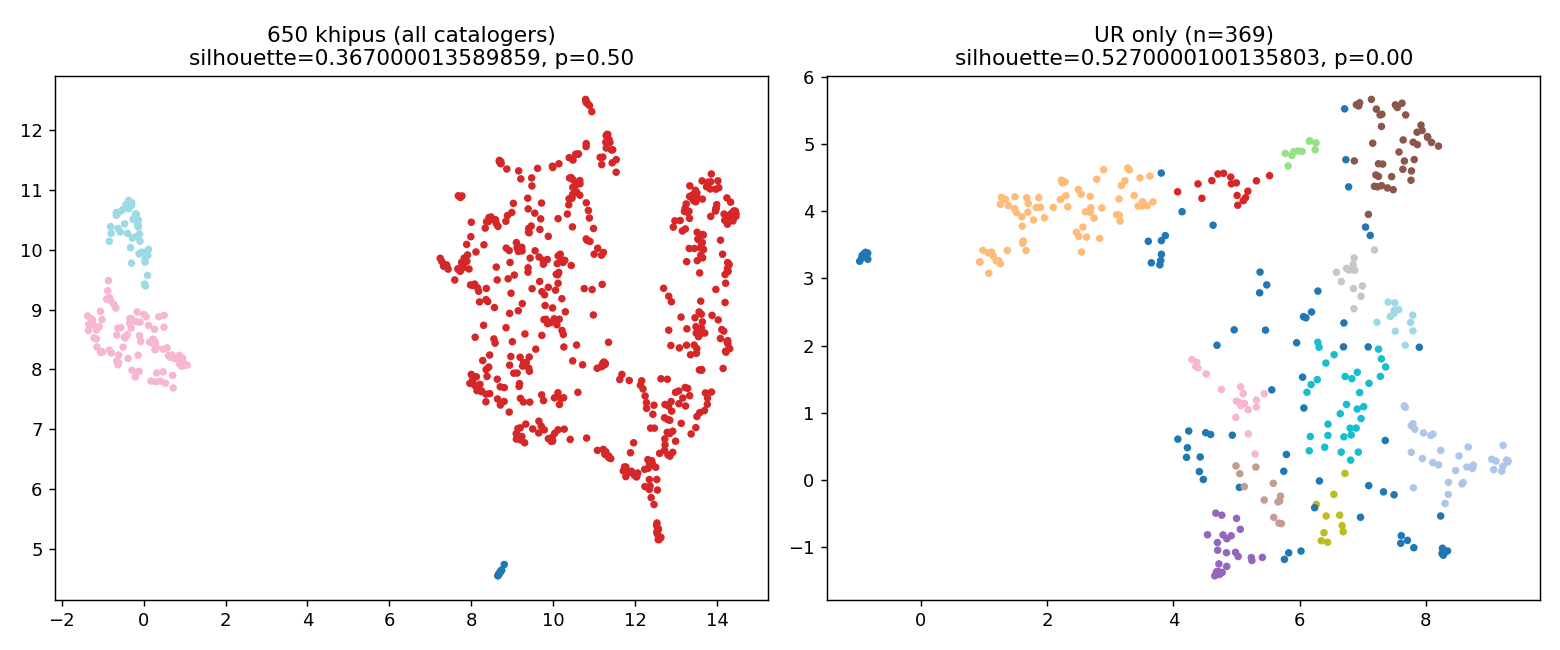}
\caption{UMAP projections of structural features, colored by HDBSCAN cluster assignment, from the primary computing environment. Left: all 650 khipus pooled across catalogers (robust negative result, reproduced on a second machine). Right: restricted to the UR cataloger ($n=369$); this panel's specific embedding and silhouette score are \emph{not} reproducible on a second machine (see text above).}
\label{fig:clustering}
\end{figure}

This negative result corroborates an independent conclusion reached by the KFG project itself: its ``Analytical Overview'' page reports that hierarchical clustering on textual khipu descriptions, color/value clustering, and image-based clustering were all tried and found unreliable for discovering khipu typologies, useful only for detecting physical duplicates---precisely the artifact this paper's Section~2 relied on to validate the 35 SQL-only khipus.

\section{Calendar Khipu Candidate Replication}
\label{sec:calendar}

Where generic structural clustering failed, a targeted, literature-grounded heuristic succeeds. The KFG's ``Percent of S and Z Knots'' analysis proposes that calendar/astronomical khipus organize S- and Z-direction knots into spatial blocks (\textit{tokapu}) along the cord, rather than distributing them uniformly, and publishes 12 candidate khipus found by this method (10 of which are present in the current 650-khipu sample).

This paper replicates that heuristic directly from the SQL \texttt{cord}/\texttt{knot} tables, restricted to level-1 (non-subsidiary) pendant cords: (1) filter to khipus with a balanced S/Z knot mix ($\geq30\%$ of each); (2) compute each knot direction's mean position (``moment'') along the cord; (3) retain khipus where the S-moment, overall moment, and Z-moment fall in a strict spatial order ($s < \text{mid} < z$ or the reverse), with each moment separated from the midpoint by more than 2 position units.

\textbf{Validation.} 8 of the 10 published candidates present in the sample are recovered (KH0063, KH0097, KH0118, KH0300, KH0311, KH0405, KH0433, KH0436). Of the 2 not recovered, KH0033 has 0\% S-knots (fails the balance filter outright) and KH0349 falls just below the 30\% threshold (25.99\% S-knots)---both plausible edge cases rather than replication failures.

\textbf{New candidates.} The same criterion, applied to the full sample, yields 42 total candidates, including the 8 recovered above and 34 not on the published list (e.g.\ UR093 ``Calliope,'' UR053C ``waterfall khipu,'' UR110, UR242). The exact additional filtering the KFG applied to narrow $\sim$69 balanced-mix khipus down to its published 12 is not available in the public materials and likely included manual visual review; the 34 new candidates here are therefore proposed, not confirmed, and require the same manual inspection before being asserted as calendar khipus.

\section{Discussion}

Three methodological lessons generalize beyond this dataset. First, in curated archival collections assembled from multiple sources over decades, the identity of the original recorder is often a stronger predictor of what gets measured than any property of the object itself; the AS/HP results here (Section~\ref{sec:ascher}) show that this confound can masquerade as a substantive finding (a cultural or regional difference) until explicitly tested with variance inflation factors. Second, apparent clustering structure must be judged against a null model, not against intuition about whether the silhouette score ``looks good'': the 650-khipu pooled analysis produced a visually clean 3-way split before the documentation confound was identified, and a properly randomized baseline was the only thing that exposed it as an artifact. Third, targeted feature engineering grounded in domain literature (Section~\ref{sec:calendar}) outperformed generic structural features (Section~\ref{sec:clustering}) by a wide margin, recovering 80\% of known positive cases where generic clustering recovered none distinguishable from noise---an argument for investing in literature-informed features before scaling up unsupervised methods on under-theorized structural data.

Substantively, the one robust positive finding---HP khipus carrying fewer distinct Ascher sum-relation types than UR khipus, independent of size and documentation---is modest but genuine, and raises a question for khipu specialists rather than answering one: whether Proyecto Quipu's cataloging protocol under-detected relations that were present, or whether the Pachacamac-heavy HP collection (57\% of its khipus) genuinely differs in accounting structure from Urton's more geographically dispersed sample. This paper's data cannot distinguish the two; disaggregating HP's Pachacamac and non-Pachacamac khipus found no difference between them (both below the UR baseline), which weakens the site-based explanation without ruling it out.

\section{Limitations}

\begin{itemize}[leftmargin=*]
\item \textbf{AS collinearity is unresolved, not disproven.} The near-total absence of side/direction documentation for Ascher-cataloged khipus (84--89\% unknown) makes its cataloger effect on sum-relation counts statistically inseparable from a documentation-completeness effect. Resolving this requires either additional AS khipus with fuller documentation or an external, non-KFG source of side/direction data for the same khipus.
\item \textbf{The 34 new calendar-khipu candidates are unvalidated.} They satisfy the quantitative S/Z spatial-block criterion but have not been checked against the khipu's graphical representation, as the KFG's original 12-candidate list was. Several (e.g.\ small khipus with few level-1 cords) may be false positives driven by small-sample moment estimates.
\item \textbf{The within-cataloger clustering result (Section~\ref{sec:clustering}) is not reproducible across runtime environments:} silhouette$=0.474$, $p=0.06$ in the primary environment versus silhouette$=0.527$, $p\approx0.00$ on a second machine running identical code, with package-version differences explicitly ruled out as the cause. It should not be treated as evidence of structure; a pre-registered replication in a fixed, containerized environment is needed before this analysis is repeated.
\item \textbf{Five structural features could not be reproduced from raw SQL} (\texttt{max\_subsidiary\_branches}, \texttt{fanout\_ratio}, \texttt{mean\_cord\_value}, \texttt{num\_banded\_clusters}, \texttt{num\_seriated\_clusters}); they are carried over from the published KFG summary table as-is, and their exact computation is not independently verifiable from the materials available here.
\item \textbf{Scope.} All findings are specific to the KFG's 650-khipu sample and its particular catalogers; they should not be read as general claims about Inka khipu-keeping practice without corroboration from khipus outside this collection.
\end{itemize}

\section{Conclusion}

This paper's contribution is a disciplined negative result and one validated positive one, arrived at by making two archival confounds explicit and testing every structural claim against null models before interpreting it. Khipu size and cataloger identity jointly explain most of what looked, at first pass, like structural variation; once controlled for, generic clustering offers nothing a random Gaussian cloud does not, while a single literature-grounded heuristic (S/Z spatial blocks) recovers most of a known positive set and extends it.

\textbf{Concrete next steps:} (1) manually review the 34 new calendar-khipu candidates against their graphical representations, prioritized by spatial-block strength (\texttt{delta\_s}); (2) seek an independent documentation source to resolve the AS collinearity; (3) extend the within-cataloger null-model test to AS and HP individually, now that both are large enough subsamples, once documentation gaps permit meaningful structural features; (4) before re-attempting within-cataloger clustering on any khipu collection, fix the full software environment (container image, pinned transitive dependencies) and verify bitwise reproducibility across at least two machines, given the cross-environment instability documented here.

\noindent\textit{All KFG web sources accessed 2026-10-28; page content may change in later KFG updates.}


\begin{thebibliography}{9}

\bibitem{contreras2026}
Contreras, M.
\textit{Structural Pattern Mining in Inka Khipus}.
arXiv:2607.00185 (2026).
Prior OKR-based preprint superseded in scope by this paper.

\bibitem{medrano2024}
Medrano, M. \& Khosla, A.
\textit{How Can Data Science Contribute to Understanding the Khipu Code?}
Companion data and code.
\url{https://github.com/khipufieldguide/data-science-khipu-article},
DOI: \url{https://doi.org/10.5281/zenodo.10472429}. MIT license.

\bibitem{brezine2023}
Brezine, C.J., Clindaniel, J., Ghezzi, I., Hyland, S. \& Medrano, M.
A New Naming Convention for Andean Khipus.
\textit{Latin American Antiquity} (2023).
DOI: \url{https://doi.org/10.1017/laq.2023.71}.

\bibitem{urton2001}
Urton, G.
A Calendrical and Demographic Tomb Text from Northern Peru.
\textit{Latin American Antiquity} 12(2), 127--147 (2001).
Cited via KFG catalog entry UR006/KH0242, \url{https://www.khipufieldguide.com/catalog/UR006.html}.

\bibitem{kfg-analytical}
Khipu Field Guide.
Analytical Overview.
\url{https://khipufieldguide.com/notebook/analyses/analytical_overview.html}.
Independent conclusion that generic clustering is unreliable for khipu typology.

\bibitem{kfg-sz}
Khipu Field Guide.
Percent of S and Z Knots.
\url{https://khipufieldguide.com/notebook/analyses/percent_z_knots.html}.
Source of the calendar-khipu spatial-block heuristic and its 12 published candidates.

\bibitem{kfg-dup}
Khipu Field Guide.
Duplicate Khipu Search.
\url{https://khipufieldguide.com/notebook/appendix/notebooks/07\%20-\%20Removing\%20Duplicate\%20Khipus.html}.
List of 10 physical-duplicate pairs used to validate the 35 SQL-only khipus in Section~2.

\end{thebibliography}
\end{document}